\documentclass{article}

\usepackage[utf8]{inputenc} 
\usepackage[T1]{fontenc}    
\usepackage{hyperref}       
\usepackage{url}            
\usepackage{booktabs}       
\usepackage{amsfonts}       
\usepackage{nicefrac}       
\usepackage{microtype}      
\usepackage[pdftex]{graphicx}
\usepackage{comment}

\title{Large-Scale Landslides Detection from Satellite Images with Incomplete Labels}

\author{%
  Masanari Kimura \\
  Ridge-i Inc.\\
  \texttt{mkimura@ridge-i.com} \\
}

\begin{document}

\maketitle

\begin{abstract}
Earthquakes and tropical cyclones cause the suffering of millions of people around the world every year.
The resulting landslides exacerbate the effects of these disasters.
Landslide detection is, therefore, a critical task for the protection of human life and livelihood in mountainous areas.
To tackle this problem, we propose a combination of satellite technology and Deep Neural Networks (DNNs).
We evaluate the performance of multiple DNN-based methods for landslide detection on actual satellite images of landslide damage.
Our analysis demonstrates the potential for a meaningful social impact in terms of disasters and rescue.
\end{abstract}

\section{Introduction}
Landslides, slope failures caused by heavy rain and earthquakes, threaten human life and property.
They are continuous and recurrent- motion may continue even after the initial landslide.
To protect those in affected areas, understanding where earthquake-triggered or rainfall-triggered landslides have occurred is critical.

Satellite imaging has attracted attention as a means of detecting topographical changes.
Satellites can regularly observe a given wide area of land, instrumental for disaster detection.
The recent development of satellite technology has drastically increased the maximum observable resolution, enabling more accurate analysis.
Many studies focus on landslide detection from satellite images~\cite{nichol2005satellite,cheng2004locating}.

Detection attempts usually rely on annotated data~\cite{ghorbanzadeh2019evaluation,amit2017disaster}.
However, accurately annotating such data when an emergency occurs, is not always easy.
Therefore, experimenting with multiple detection methods and evaluating their performance, according to the presence or absence of labels, is essential. 
In this paper, we perform such an evaluation using several DNN-based landslide detection methods.

\section{Supervised Landslides Detection from Satellite Images}

\begin{figure*}
  \centering
  \includegraphics[scale=0.4]{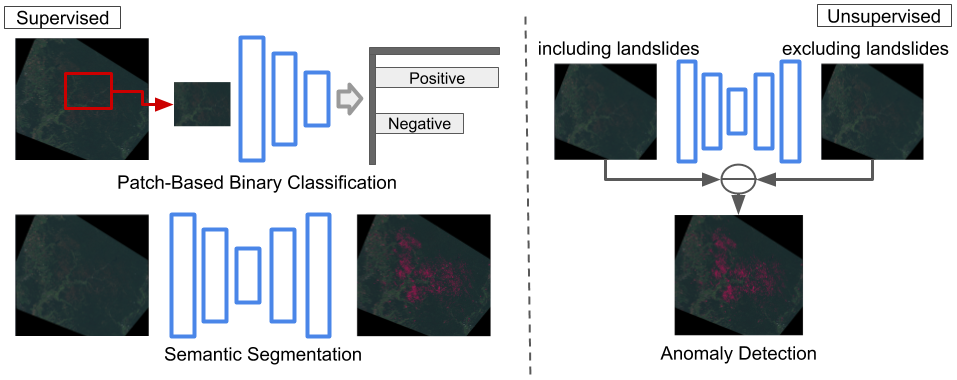}
  \caption{Overview of landslides detection methods.
  The left side is the supervised methods, and the right side is the unsupervised method.
  \label{fig:overview_methods}}
\end{figure*}

We applied landslide detection methods based on multiple DNNs and compared them.
In this section, we introduce several DNN-based methods for landslide detection in satellite images.
We give an overview of each method in the left of Figure~\ref{fig:overview_methods}.

\subsection{Patch-Based Binary Classification}
In Patch-Based Binary Classification, we divide the input image into patches of a certain size and classify the presence of landslides for each patch. We use this approach as a baseline, as it is a very orthodox convolutional neural network based approach.
We adopted the ResNet50~\cite{he2016deep} as the architecture.

\subsection{Semantic Segmentation}
We applied DeepLabv3+~\cite{chen2018encoder}, a pixel-by-pixel classification method.
This is one of the most successful semantic segmentation models.

With the disadvantage of higher annotation costs compared to the baseline, it allows more precise predictions.
We need to select the method appropriately considering the trade-off between annotation cost and detection accuracy.

\section{Unsupervised Landslides Detection via Deep Anomaly Detection}
Supervised landslide detection is powerful, but has several drawbacks: (1) Annotation is required. Such annotations require costly domain expertise. (2) Label imbalance.
In general, since disasters are rare, so are contemporary labels.

To tackle this problem, we propose an unsupervised anomaly detection method.
There are several such methods using DNNs~\cite{schlegl2017unsupervised,kimura2018anomaly,kimura2019pnunet}.

In this paper, we use the U-Net architecture~\cite{ronneberger2015u} for anomaly detection.
We give an overview of unsupervised anomaly detection methods in right of Figure~\ref{fig:overview_methods}.

We assume the noise $z$ to be sampled from the distribution $\mathcal{P}(z)$ in the input image $x$.
The noise removal component $f(\cdot)$ removes this noise from the image. 
The loss function is as follows:
\begin{equation}
    \mathcal{L} = \mathcal{L}_r(x, f_r(x + z)),\ \ z\sim\mathcal{P}(z),
\end{equation}
Here, we use Structural SIMilarity (SSIM) as the reconstruction error $\mathcal{L}_r$ and $\mathcal{P}(z)$ = $\mathcal{N}(0, 1)$.
The proposed method is trained to remove random noise sampled from the normal distribution and behaves to remove actual abnormal areas during inference.

\section{Experimental Results}

\begin{figure*}
  \centering
  \includegraphics[scale=0.44]{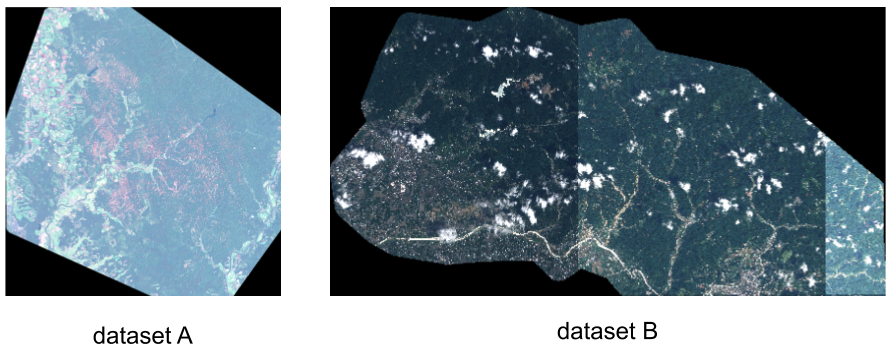}
  \caption{Overview of datasets.
  Here, dataset $A$ is completely labeled by experts, while dataset $B$ has only a small amount of labels for evaluation.
  Therefore, dataset $B$ is not used for quantitative evaluation, but only for visualization experiment.
  \label{fig:dataset_overview}}
\end{figure*}

\begin{figure*}
  \centering
  \includegraphics[scale=0.47]{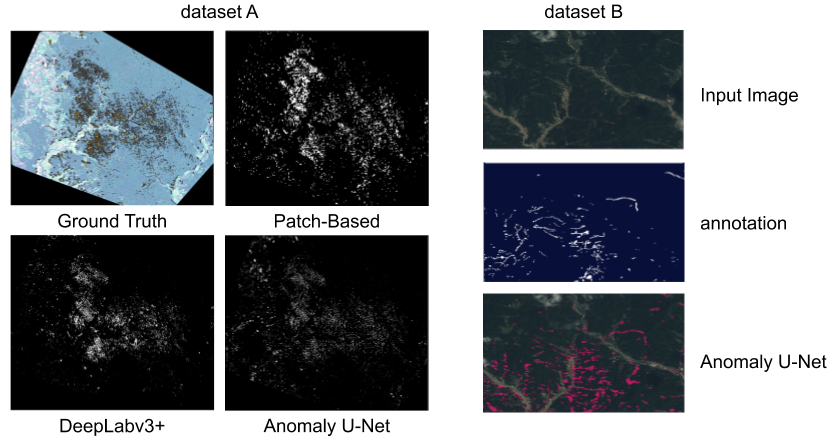}
  \caption{Qualitative evaluation for both dataset $A$ and dataset $B$.
  The landslide area in dataset $A$ is visualized using several methods.
  On the other hand, because dataset $B$ lacks label information, we visualized it using only the anomaly detection method.
  \label{fig:qualitative_evaluation}}
\end{figure*}

In the experiments, we use two datasets~\ref{fig:dataset_overview}.
Dataset $A$ contains complete annotations, and an image size of $20000\times 18957$.
Dataset $B$ has incomplete labels, and an image size of $43008\times 36864$.
In supervised methods, we used a 50:50 training-to-test data ratio.
In the unsupervised method, we adapted as training data those images with less than 50\% of their area exhibiting landslides.




Figure~\ref{fig:qualitative_evaluation} shows the qualitative evaluation for both dataset $A$ and dataset $B$.
The annotations of dataset $B$, as shown in Figure~\ref{fig:qualitative_evaluation}, are applied only to some areas, based on expert qualitative evaluation. 
We can see that each method enables landslide visualization, providing at least a rough grasp of the damage.

\begin{table}[t]
\centering
\caption{Qualitative evaluation of patch-based binary classifier for dataset $A$.
This shows that performance is sufficient to understand where landslides are occurring.
\label{table:hokkaido_patch_qual}}
\begin{tabular}{l|cccc} \hline
network (dataset) & Accuracy (\%) & F1 Score & Precision & Recall \\
\hline\hline
ResNet50 (train)  & 93.5          & 0.663    & 0.572     & 0.788  \\
ResNet50 (test)   & 89.0          & 0.660    & 0.542     & 0.846  \\ \hline
\end{tabular}
\end{table}

\begin{table}[t]
\centering
\caption{Qualitative evaluation of semantic DeepLabv3+ and Anomaly U-Net for dataset $A$.
DeepLabv3+ can detect landslide areas with high accuracy, but require meticulous annotation.
Despite these accuracy limitations however, the Anomaly U-Net permits a rough damage assessment without label information.
\label{table:hokkaido_segmentation_anomaly}}
\begin{tabular}{l|cc} \hline
Method        & Mean IOU & Annotation \\ \hline\hline
DeepLabv3+~\cite{chen2018encoder}    & 0.8105   & Supervised \\
Anomaly U-Net & 0.7102   & Unsupervised \\ \hline
\end{tabular}
\end{table}

Table~\ref{table:hokkaido_patch_qual} shows the quantitative evaluation of the patch-based binary classifier, for dataset $A$.
This shows that performance is sufficient to understand where landslides are occurring.
On the other hand, the patch-based method divides the input image into rough grids, preventing a detailed understanding of the damage.
In order to handle a more accurate landslide area, it is necessary to consider another method.

Table~\ref{table:hokkaido_segmentation_anomaly} shows the experimental results for dataset $A$, using semantic segmentation and anomaly detection methods.
Semantic segmentation techniques can detect landslide areas with high accuracy, but require meticulous annotation.
Despite these accuracy limitations, however, the anomaly detection method permits a rough damage assessment without label information.

\section{Conclusion and Discussion}
In order to prepare for emergencies, understanding the performance of each method, based on whether or not data can be labeled, is critical.
In this paper, we experiment and analyze the landslide detection task using multiple methods based on DNNs.
We concluded that the choice of detection method depends highly on the availability of labels.
When labels are absent, we propose to use the unsupervised anomaly detection model to achieve landslide detection.
This report makes it possible to quickly understand the compatibility between each detection method and the exact circumstances of future disasters.

\section*{Acknowledgement}
This research was supported by Japan Aerospace eXploration Agency~(JAXA). I would like to thank my colleague Aaron Bell for comments that greatly improved the manuscript.
I also thank my boss Takashi Yanagihara and Issei Sugiyama for manage overall this project.

\bibliographystyle{unsrt}

\end{document}